\documentclass[11pt,letterpaper]{article}
\usepackage{emnlp2016}
\usepackage{times}
\usepackage{latexsym}
\usepackage{amsmath}

\emnlpfinalcopy

\def\emnlppaperid{298}

\usepackage{examples}
\usepackage{balance}
\usepackage{cgloss4e}
\usepackage{verbatim}
\usepackage{cprotect}
\usepackage{listings}
\usepackage{url}
\usepackage{amsmath}
\usepackage{stackrel}
\usepackage[pdftex]{graphicx}
\usepackage{dblfloatfix}
\usepackage{graphics}
\usepackage{array}
\newcolumntype{x}[1]{>{\centering\arraybackslash\hspace{0pt}}p{#1}}
\newcolumntype{y}[1]{<{\hspace{-.15cm}}p{#1}}
\usepackage[usenames,dvipsnames,svgnames,table]{xcolor}

\usepackage{amssymb,amsmath,epsfig}
\usepackage{mathpartir}
\usepackage{proof}
\usepackage{amsthm}
\usepackage{xspace}
\usepackage{algorithm}
\usepackage[noend]{algpseudocode}
\algrenewcommand{\algorithmicindent}{1em}

\usepackage{enumitem}

\newcommand{\notes}[1]{}

 \theoremstyle{definition}
 
\theoremstyle{plain}

\newcommand{\ith}[1]{\ensuremath{i^{{th}}}}

\newcount\permx
\newcount\permy
\def\permdot#1#2{
\permx=#1 \advance\permx by-1
\permy=#2 \advance\permy by-1
\psframe[fillcolor=black, fillstyle=solid]
(\permx,\permy)(#1, #2)
}

\newcommand{\boxnum}[1]{{\setlength{\fboxsep}{1pt}\raisebox{1pt}{\hspace{1pt}\fbox{\tiny #1}\hspace{1pt}}}}
\newcommand{\ind}[1]{\ensuremath{_{\kern-0.5pt\boxnum{#1}}}}

\def\namecite{\newcite}

\newcommand{\smallnt}[1]{\ensuremath{_{\mbox{\tiny PP}}}\xspace}

\newcommand{\pseudocode}{Algorithm}
\floatname{algorithm}{\pseudocode}

\iffalse

\else

\fi

\newcommand{\ter}{{\sc Ter}\xspace}
\newcommand{\bp}{{\sc Bp}\xspace}
\newcommand{\tb}{{\sc T-b}\xspace}
\newcommand{\bleu}{{\sc Bleu}\xspace}

\usepackage{multirow}
\usepackage{tikz}
\usetikzlibrary{positioning}
\usepackage{tikz-qtree}
\usetikzlibrary{arrows}
\title{Supervised Attentions for Neural Machine Translation}

\iftrue
\author{
Haitao Mi \;\;  \;\; Zhiguo Wang \;\;  \;\; Abe Ittycheriah\\
T.J.~Watson Research Center \\
IBM \\
1101 Kitchawan Rd, Yorktown Heights, NY 10598 \\
{\tt \{hmi, zhigwang, abei\}@us.ibm.com}
}
\fi
\date{}

\begin{document}

\maketitle

\begin{abstract}
In this paper, we improve the attention or alignment accuracy of neural machine translation
by utilizing the alignments of training sentence pairs.
We simply compute the distance between the machine attentions and the ``true'' alignments,
and minimize this cost in the training procedure. 
Our experiments on large-scale 
Chinese-to-English task show that our model improves both translation and alignment 
qualities significantly over the large-vocabulary neural machine translation system,
and even beats a state-of-the-art traditional syntax-based system.
\end{abstract}

\section{Introduction}
\label{sec:intro}
Neural machine translation (NMT) has gained popularity in recent two years~\cite{bahdanau+:2014,jean+:2015,luong+:2015}, 
especially for the attention-based models of \namecite{bahdanau+:2014}.

The attention model plays a crucial role in NMT, as it shows which source word(s) the 
model should focus on in order to predict the next target word.
However, the attention or alignment quality of NMT is still very low~\cite{mi+cov:2016,tu+:2016}.

In this paper, we alleviate the above issue by utilizing the {\em alignments} 
(human annotated data or machine alignments) of the training set.
Given the alignments of all the training sentence pairs, 
we add an alignment distance cost to the objective function.
Thus, we not only maximize the log translation probabilities,
but also minimize the alignment distance cost.
Large-scale experiments over Chinese-to-English on 
various test sets show that
our best method for a single system improves the translation quality significantly 
over the large vocabulary NMT system (Section~\ref{sec:exps}) 
and beats the state-of-the-art syntax-based system.

\section{Neural Machine Translation}
\label{sec:nmt}
\begin{figure*}[!t]
\centering
\includegraphics[width=0.85\textwidth]{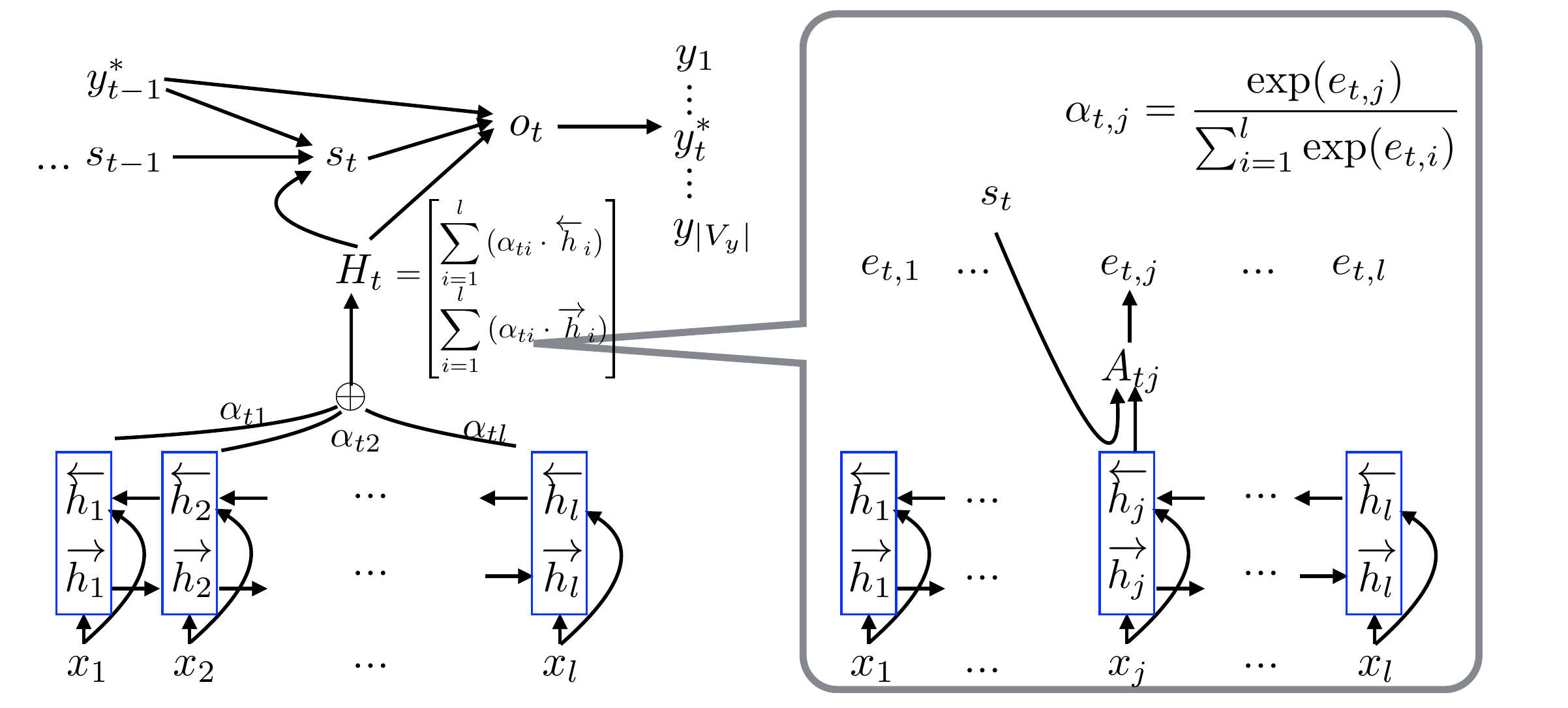}
\caption{The architecture of attention-based NMT \protect\cite{bahdanau+:2014}.
The source sentence ${\bf x}=(x_1, ... , x_l)$ with length $l$, 
$x_l$ is an end-of-sentence token $\langle\text{eos}\rangle$ on the source side.
The reference translation is ${\bf y^*}=(y^*_1, ... , y^*_m)$ with length $m$,
similarly, $y^*_m$ is the target side $\langle\text{eos}\rangle$.
$\protect\overleftarrow{h_i}$ and $\protect\overrightarrow{h_i}$ are bi-directional encoder states. 
$\alpha_{t,j}$ is the attention probability at time $t$, position $j$. 
$H_{t}$ is the weighted sum of encoding states. $s_t$ is a hidden state. 
$o_t$ is an output state. 
Another one layer neural network projects $o_t$ to the target output vocabulary, 
and conducts softmax to predict the probability distribution over the output vocabulary. 
The attention model (the right box) is a two layer feedforward neural network, 
$A_{t,j}$ is an intermediate state, then another layer converts it into a real number $e_{t,j}$, 
the final attention probability at position $j$ is $\alpha_{t,j}$.}\label{fig:att}
\end{figure*}

As shown in Figure~\ref{fig:att}, 
attention-based NMT \cite{bahdanau+:2014} 
is an encoder-decoder network.  
the encoder employs a bi-directional recurrent neural network to 
encode the source sentence ${\bf{x}}=({x_1, ... , x_l})$, 
where $l$ is the sentence length (including the end-of-sentence $\langle\text{eos}\rangle$), into
a sequence of hidden states ${\bf{h}}=({h_1, ..., h_l})$,
each $h_i$ is a concatenation of a left-to-right $\overrightarrow{h_i}$
and a right-to-left $\overleftarrow{h_i}$.

Given ${\bf h}$, the decoder predicts the target translation
by maximizing the conditional log-probability of the 
correct translation ${\bf y^*} = (y^*_1, ... y^*_m)$, where 
$m$ is the sentence length (including the end-of-sentence). 
At each time $t$, the probability of each word $y_t$ from a target vocabulary $V_y$ is:
\begin{equation}
\label{eq:py}
p(y_t|{\bf h}, y^*_{t-1}..y^*_1) = g(s_t, y^*_{t-1}),
\end{equation}
where $g$ is 
a two layer feed-forward neural network 
over the embedding of the previous word $y^*_{t-1}$, and  
the hidden state $s_t$. 
The $s_t$ is computed as:
\begin{equation}
s_t = q(s_{t-1}, y^*_{t-1}, H_{t})
\end{equation}
\begin{equation}
H_t = 
\begin{bmatrix}
\sum_{i=1}^{l}{(\alpha_{t,i} \cdot \overleftarrow{h}_i)} \\
\sum_{i=1}^{l}{(\alpha_{t,i} \cdot \overrightarrow{h}_i)} \\
\end{bmatrix},
\end{equation}
where $q$ is a gated recurrent units, $H_t$ is a weighted sum of ${\bf h}$;
the weights, $\alpha$, are computed with a two layer feed-forward neural network $r$:
\begin{equation}
\alpha_{t,i} = \frac{\exp\{r(s_{t-1}, h_{i}, y^*_{t-1})\}}{\sum_{k=1}^{l}{\exp\{r(s_{t-1}, h_{k}, y^*_{t-1})\}}}
\end{equation}
We put all $\alpha_{t,i}$ ($t = 1...m$, $i= 1...l$) into a matrix $\mathcal{A'}$, 
we have a matrix (alignment) like (c) in Figure~\ref{fig:alignment}, 
where each row (for each target word) is a probability distribution over the source sentence ${\bf x}$.

The training objective is to maximize the conditional log-probability of 
the correct translation $y^*$ given $x$ with respect to the parameters $\theta$
\begin{equation}
\theta^* = \arg\max_{\theta}{\sum_{n=1}^{N}{\sum_{t=1}^{m}{\log p(y^{*n}_t|{\bf x}^n, y^{*n}_{t-1}..y^{*n}_1)}}},
\label{eq:nmt}
\end{equation}
where $n$ is the $n$-th sentence pair $({\bf x}^n, {\bf y^*}^n)$ in the training set, $N$ is the total number of pairs.

\section{Alignment Component}
\label{sec:alignment}

\begin{figure*}[!t]
\centering
\includegraphics[width=0.9\textwidth]{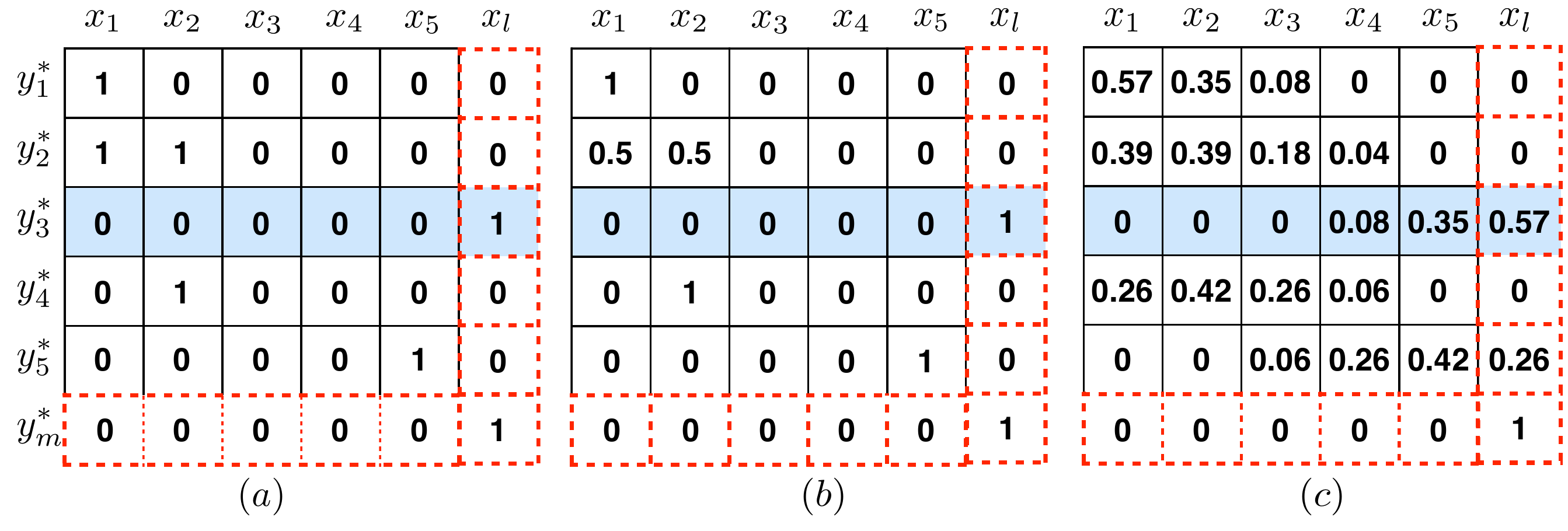}
\caption{Alignment transformation. 
A special token, $\langle\text{eos}\rangle$, is introduced to the source sentence, 
we align all the unaligned target words ($y^*_3$ in this case) to $\langle\text{eos}\rangle$.
(a): the original alignment matrix $\mathcal{A}$ from GIZA++ or MaxEnt aligner.
(b): simple normalization by rows (probability distribution over the source sentence $\bf {x}$). 
(c): smoothed transformation followed by normalization by rows, and typically, 
we always align end-of-source-sentence $x_l$ to end-of-target-sentence $y_m$ by probability one.
\label{fig:alignment}}
\end{figure*}

The attentions, $\alpha_{t,1} ... \alpha_{t,l}$, in each step $t$ play an important role in NMT.
However, the accuracy is still far behind the traditional MaxEnt alignment model in terms of 
alignment F1 score~\cite{mi+:2016,tu+:2016}.
Thus, in this section, we explicitly add an alignment distance to the objective function in Eq.~\ref{eq:nmt}.
The ``truth'' alignments for each sentence pair can be from human annotated data,
unsupervised or supervised alignments (e.g. GIZA++~\cite{och+ney:2000} or MaxEnt~\cite{abe+salim:2005}).

Given an alignment matrix $\mathcal{A}$ for a sentence pair (${\bf x, y}$) 
in Figure~\ref{fig:alignment} (a), where we have an end-of-source-sentence token $\langle\text{eos}\rangle = x_l$,
and we align all the unaligned target words ($y^*_3$ in this example) to $\langle\text{eos}\rangle$,
also we force $y^*_m$ (end-of-target-sentence) to be aligned to $x_l$ with probability one.
Then we conduct two transformations to get 
the probability distribution matrices ((b) and (c) in Figure~\ref{fig:alignment}).

\subsection{Simple Transformation}
The first transformation simply normalizes each row. Figure~\ref{fig:alignment} (b) shows the result matrix $\mathcal{A^*}$.
The last column in red dashed lines shows the alignments of the special end-of-sentence token $\langle\text{eos}\rangle$.

\subsection{Smoothed Transformation}
\label{sec:smooth}
Given the original alignment matrix $\mathcal{A}$, 
we create a matrix $\mathcal{A^*}$ with all points initialized with zero.
Then, for each alignment point $\mathcal{A}_{t,i} = 1$,
we update $\mathcal{A^*}$ by adding a Gaussian distribution, $g(\mu, \sigma)$, with a window size $w$ ($t$-$w$, ... $t$ ... $t$+$w$).
Take the $\mathcal{A}_{1, 1} = 1$ for example, we have $\mathcal{A^*}_{1, 1}$ += $1$, $\mathcal{A^*}_{1, 2}$ += $0.61$,
and $\mathcal{A^*}_{1, 3}$ += $0.14$ with $w$=2, $g(\mu, \sigma)$=$g(0, 1)$.
Then we normalize each row and get (c). In our experiments, we use a shape distribution, where $\sigma$ = $0.5$.

\subsection{Objectives}
{\bf Alignment Objective}: 
Given the ``true'' alignment $\mathcal{A^*}$, 
and the machine attentions $\mathcal{A'}$ produced by NMT model, 
we compute the Euclidean distance bewteen $\mathcal{A^*}$ and $\mathcal{A'}$.
\begin{equation}
d(\mathcal{A'}, \mathcal{A^*}) = \sqrt{\sum_{t=1}^{m}{\sum_{i=1}^{l}{(\mathcal{A'}_{t,i} - \mathcal{A^*}_{t,i})^2}}}.
\label{eq:alignment}
\end{equation}
{\bf NMT Objective}: We plug Eq.~\ref{eq:alignment} to Eq.~\ref{eq:nmt}, we have
\begin{equation}
\begin{split}
\theta^* = \arg\max_{\theta}{\sum_{n=1}^{N}} & {\Bigg\{\sum_{t=1}^{m}{\log p(y^{*n}_t|{\bf x}^n, y^{*n}_{t-1}..y^{*n}_1})} \\
     & - d(\mathcal{A'}^n, \mathcal{A^*}^n) \Bigg\}.
\end{split}
\label{eq:mix}
\end{equation}
There are two parts: translation and alignment, 
so we can optimize them jointly, or separately (e.g. we first optimize alignment only, then optimize translation).
Thus, we divide the network in Figure~\ref{fig:att} into alignment {\bf A} and translation {\bf T} parts: 
\begin{enumerate}
\item[$\bullet$] {\bf A}: all networks before the hidden state $s_t$,
\item[$\bullet$] {\bf T}: the network $g(s_t, y^*_{t-1})$.
\end{enumerate}
If we only optimize {\bf A}, we keep the parameters in {\bf T} unchanged. 
We can also optimize them jointly {\bf J}.
In our experiments, we test different optimization strategies.

\section{Related Work}
\label{sec:related}
In order to improve the attention or alignment accuracy, 
\namecite{cheng+:2016} adapted the agreement-based learning~\cite{liang+:2006,liang+:2008},
and introduced a combined objective that takes into account 
both translation directions (source-to-target and target-to-source)
and an agreement term between the two alignment directions.
By contrast, our approach directly uses and optimizes NMT parameters 
using the ``supervised'' alignments.

\section{Experiments}
\label{sec:exps}
\begin{table*}[t]
\centering
\tabcolsep=0.12cm
\begin{tabular}{c|c|c||ccc|ccc|ccc||c}
\multicolumn{3}{c||}{}               & \multicolumn{3}{c|}{\multirow{2}{*}{MT06}} & \multicolumn{6}{c||}{MT08} & \multirow{2}{*}{avg.}\\
\multicolumn{3}{c||}{single system}  & & & & \multicolumn{3}{c|}{News} & \multicolumn{3}{c||}{Web}             &     \\\cline{4-13}
\multicolumn{3}{c||}{}               & \bp & \bleu & \tb & \bp & \bleu & \tb & \bp & \bleu & \tb  & \tb \\ 
\hline
\multicolumn{3}{c||}{Tree-to-string} & 0.95 & 34.93 & 9.45 & 0.94 & 31.12 & 12.90 & 0.90 & 23.45 & 17.72 & 13.36 \\
\hline \hline
\multicolumn{3}{c||}{Cov. LVNMT {\small \cite{mi+:2016}}} & 0.92 & 35.59 & 10.71 & 0.89 & 30.18 & 15.33 & 0.97 & 27.48 & 16.67 & 14.24 \\
\hline
\parbox[t]{3mm}{\multirow{7}{*}{\rotatebox[origin=c]{90}{+Alignment}}} & \multirow{4}{*}{\bf Zh $\rightarrow$ En} & {\bf A} $\rightarrow$ {\bf J} & 0.95 & 35.71 & 10.38 &  0.93 & 30.73 & 14.98 & 0.96 & 27.38 & 16.24 & 13.87 \\
 & & {\bf A} $\rightarrow$ {\bf T}     & 0.95 & 28.59 & 16.99 & 0.92 & 24.09 & 20.89 & 0.97 & 20.48 & 23.31 & 20.40 \\
 & & {\bf A} $\rightarrow$ {\bf T} $\rightarrow$ {\bf J} & 0.95 & 35.95 & 10.24 & 0.92 & 30.95 & 14.62 & 0.97 & 26.76 & 17.04 & 13.97 \\
 & & {\bf J}                    & 0.96 & 36.76 &  9.67 & 0.94 & 31.24 & 14.80 & 0.96 & 28.35 & {\bf 15.61} & 13.36 \\\cline{2-13}
 & {\bf GDFA} & {\bf J}               & 0.96 & 36.44 & 10.16 & 0.94 & 30.66 & 15.01 & 0.96 & 26.67 & 16.72       & 13.96 \\\cline{2-13}
 & \multirow{2}{*}{\bf MaxEnt}& {\bf J} & 0.95 & 36.80 &  {\bf 9.49} & 0.93 & 31.74 & 14.02 & 0.96 & 27.53 & 16.21 & 13.24  \\\cline{3-13}
 & & {\bf J + Gau.} & 0.96 & {\bf 36.95} &  9.71 & 0.94 & {\bf 32.43} & {\bf 13.61} & 0.97 & {\bf 28.63} & 15.80 & {\bf 13.04} \\
\hline
\end{tabular}
\caption{Single system results in terms of (\ter-\bleu)/2 (\tb, the lower the better) 
on 5 million Chinese to English training set. 
\bp denotes the brevity penalty.
NMT results are on a large vocabulary ($300k$) and with UNK replaced. 
The second column shows different alignments ({\bf Zh $\rightarrow$ En} (one direction),
{\bf GDFA} (``grow-diag-final-and''), and {\bf MaxEnt}~\protect\cite{abe+salim:2005}. 
{\bf A}, {\bf T}, and {\bf J} mean optimize alignment only, translation only, and jointly.
{\bf Gau.} denotes the smoothed transformation.
\label{tab:zhen}}
\end{table*}

\subsection{Data Preparation}
We run our experiments on Chinese to English task.
The training corpus consists of approximately 5 million 
sentences available within the DARPA BOLT Chinese-English task. 
The corpus includes a mix of newswire, broadcast news, and webblog.
We do not include HK Law, HK Hansard and UN data. 
The Chinese text is segmented with a segmenter trained on CTB data 
using conditional random fields (CRF). 
Our development set is the concatenation of several tuning sets 
(GALE Dev, P1R6 Dev, and Dev 12)
initially released under the DARPA GALE program.
The development set is 4491 sentences in total.
Our test sets are NIST MT06 
(1664 sentences)
, MT08 news (691 sentences), 
and MT08 web (666 sentences).

For all NMT systems, the full vocabulary size of the training set is $300k$.
In the training procedure, we use AdaDelta~\cite{adadelta} to 
update model parameters with a mini-batch size 80.
Following \namecite{mi+cov:2016},
the output vocabulary for each mini-batch or sentence is a sub-set of the full vocabulary.
For each source sentence, the sentence-level target vocabularies are
union of top $2k$ most frequent target words and the top 10 candidates of the word-to-word/phrase translation tables
learned from `fast\_align'~\cite{dyer+:2013}. 
The maximum length of a source phrase is 4.
In the training time, we add the reference in order to make the translation reachable.

The Cov. LVNMT system is a re-implementation of the enhanced NMT system of \namecite{mi+cov:2016}, 
which employs a coverage embedding model and achieves better performance over 
large vocabulary NMT \namecite{jean+:2015}. The coverage embedding dimension of each source word is 100.

Following \namecite{jean+:2015}, 
we dump the alignments, attentions, for each sentence, and 
replace UNKs with the word-to-word translation model or the aligned source word. 

Our traditional SMT system is a hybrid syntax-based tree-to-string model~\cite{zhao+yaser:2008},
a simplified version of the joint decoding~\cite{liu+:2009,cmejrek+:2013}.
We parse the Chinese side with Berkeley parser, and align the bilingual sentences with GIZA++ and MaxEnt.
and extract Hiero and tree-to-string rules on the training set.
Our language models are trained on the English side of the parallel corpus, 
and on monolingual corpora (around 10 billion words from Gigaword (LDC2011T07).
We tune our system with PRO~\cite{hopkins+may:2011}
to minimize (\ter - \bleu)/2~\footnote{The metric used for optimization in this work is (\ter-\bleu)/2 to prevent 
the system from using sentence length alone to impact \bleu or \ter.  
Typical SMT systems use target word count as a feature and it has been observed that \bleu can be
optimized by tweaking the weighting of the target word count with no improvement in human assessments of translation quality. 
Conversely, in order to optimize \ter shorter sentences can be produced.  
Optimizing the combination of metrics alleviates this effect~\cite{tb:2008}.} 
on the development set.

\subsection{Translation Results}
Table~\ref{tab:zhen} shows the translation results of all systems.
The syntax-based statistical machine translation model achieves an average (\ter-\bleu)/2 of 13.36 on 
three test sets. 
The Cov. LVNMT system achieves an average (\ter-\bleu)/2 of 14.24, which is about 0.9 points worse 
than Tree-to-string SMT system. 
Please note that all systems are single systems. It is highly possible that 
ensemble of NMT systems with different random seeds can lead to better results 
over SMT.

We test three different alignments: 
\begin{enumerate}
\item[$\bullet$] {\bf Zh $\rightarrow$ En} (one direction of GIZA++),
\item[$\bullet$]{\bf GDFA} (the ``grow-diag-final-and'' heuristic merge of both directions of GIZA++),
\item[$\bullet$]{\bf MaxEnt} (trained on $67k$ hand-aligned sentences). 
\end{enumerate}
The alignment quality improves from {\bf Zh $\rightarrow$ En} to {\bf MaxEnt}.
We also test different optimization strategies: {\bf J} (jointly), 
{\bf A} (alignment only), and {\bf T} (translation model only).
A combination, {\bf A} $\rightarrow$ {\bf T}, shows that we optimize {\bf A} only first,
then we fix {\bf A} and only update {\bf T} part.
{\bf Gau.} denotes the smoothed transformation (Section~\ref{sec:smooth}). 
Only the last row uses the smoothed transformation, all others use
the simple transformation.

Experimental results in Table~\ref{tab:zhen} show some interesting results.
First, with the same alignment, {\bf J} joint optimization works best than 
other optimization strategies (lines 3 to 6).
Unfortunately, breaking down the network into two separate parts ({\bf A} and {\bf T}) 
and optimizing them separately do not help (lines 3 to 5). 
We have to conduct joint optimization {\bf J} in order to get a 
comparable or better result (lines 3, 5 and 6) over the baseline system.

Second, when we change the training alignment seeds 
({\bf Zh $\rightarrow$ En}, {\bf GDFA}, and {\bf MaxEnt})
NMT model does not yield significant different results (lines 6 to 8).

Third, the smoothed transformation ({\bf J + Gau.}) gives 
some improvements over the simple transformation (the last two lines),
and achieves the best result (1.2 better than LVNMT, and 0.3 better than Tree-to-string).
In terms of \bleu scores, 
we conduct the statistical significance tests with the sign-test of~\namecite{collins+:2005}, 
the results show that the improvements of our {\bf J + Gau.} over LVNMT are significant on three test sets ($p < 0.01$).

At last, the brevity penalty (BP) consistently gets better after we add the alignment cost to NMT objective.
Our alignment objective adjusts the translation length to be more in line with the human references accordingly.

\begin{table}
\centering
\tabcolsep=0.1cm
\begin{tabular}{c|c|c|c|c|c}
\multicolumn{3}{c|}{system}    & pre. & rec. & F1 \\
\hline
\multicolumn{3}{c|}{MaxEnt}        & 74.86 & 77.10 & 75.96 \\
\hline \hline
\multicolumn{3}{c|}{Cov LVNMT {\small \cite{mi+:2016}}} & 51.11 & 41.42 & 45.76 \\
\hline
\parbox[t]{3mm}{\multirow{8}{*}{\rotatebox[origin=c]{90}{+Alignment}}} & \multirow{4}{*}{\bf Zh $\rightarrow$ En} & {\bf A}                           & 50.88 & 45.19 & 47.87 \\
 & & {\bf A $\rightarrow$ J }         & 53.18 & 49.37 & 51.21 \\
 & & {\bf A $\rightarrow$ T }         & 50.29 & 44.90 & 47.44 \\
 & & {\bf A $\rightarrow$ T $\rightarrow$ J } & 53.71 & 49.33 & 51.43 \\
 & & {\bf J}                          & {\bf 54.29} & 48.02 & 50.97 \\\cline{2-6}
 & {\bf GDFA} & {\bf J}               & 53.88 & 48.25 & 50.91 \\\cline{2-6}
 & \multirow{2}{*}{\bf MaxEnt}& {\bf J}   & 44.42 & 55.25 & 49.25 \\
 & & {\bf J + Gau.}                   & 48.90 & {\bf 55.38} & {\bf 51.94} \\
\hline
\end{tabular}
\caption{Alignment F1 scores of different models. \label{tab:align}}
\end{table}

\subsection{Alignment Results}

Table~\ref{tab:align} shows the alignment
F1 scores on the alignment test set (447 hand aligned sentences).
The MaxEnt model is trained on $67k$ hand-aligned sentences, and achieves an F1 score of 75.96.
For NMT systems, we dump the alignment matrixes and convert them into alignments 
with following steps.
For each target word, 
we sort the alphas
and add the max probability link if it is higher than 0.2.
If we only tune the alignment component ({\bf A} in line 3), 
we improve the alignment F1 score from 45.76 to 47.87. 
And we further boost the score to 50.97 by tuning alignment and translation jointly 
({\bf J} in line 7). 
Interestingly, the system using {\bf MaxEnt}
produces more alignments in the output, and results in a higher recall. 
This suggests that using {\bf MaxEnt} can lead to a sharper attention distribution,
as we pick the alignment links based on the probabilities of attentions, 
the sharper the distribution is, more links we can pick.
We believe that a sharp attention distribution is a great property of NMT.

Again, the best result is {\bf J + Gau.} in the last row, 
which significantly improves the F1 by  5 points 
over the baseline Cov. LVNMT system. 
When we use {\bf MaxEnt} alignments, {\bf J + Gau.} smoothing gives 
us about 1.7 points gain over {\bf J} system. So it looks interesting
to run another {\bf J + Gau.} over {\bf GDFA} alignment.

Together with the results in Table~\ref{tab:zhen},
we conclude that adding the alignment cost to the training objective 
helps both translation and alignment significantly.

\section{Conclusion}
In this paper, we utilize the ``supervised'' alignments, 
and put the alignment cost to the NMT objective function.
In this way, we directly optimize the attention model in a supervised way.
Experiments show significant improvements in both translation and alignment
tasks over a very strong LVNMT system.

\section*{Acknowledgment}
We thank the anonymous reviewers for useful comments.

\balance
\bibliography{emnlp2016}
\bibliographystyle{emnlp2016}

\end{document}